\documentclass[10pt,twocolumn,letterpaper]{article}
\usepackage[pagenumbers]{cvpr}
\usepackage{cvpr}              

\usepackage{graphicx}
\usepackage{amsmath}
\usepackage{amssymb}
\usepackage{booktabs}
\usepackage{setspace}
\usepackage{cuted}
\usepackage[pagebackref,breaklinks,colorlinks]{hyperref}

\usepackage[capitalize]{cleveref}
\crefname{section}{Sec.}{Secs.}
\Crefname{section}{Section}{Sections}
\Crefname{table}{Table}{Tables}
\crefname{table}{Tab.}{Tabs.}

\def\cvprPaperID{*****} 
\def\confName{CVPR}
\def\confYear{2022}

\usepackage{xcolor}
\usepackage{comment}
\usepackage{cuted}
\usepackage{hyperref}
\usepackage{cleveref}
\usepackage{placeins} 

\begin{document}
\title{KM-Speaker: Keypoint-Based Style Control for High-Quality Speech-Driven 3D Facial Animation and Dialogue Localization}
\author
{\parbox{\textwidth}{\centering Arthur Josi $^{1, 2}$\;\;
                                Emeline Got$^2$\;\;
                                Abdallah Dib$^{2}$\;\;
                                Luiz Gustavo Hafemann$^{2}$\thanks{This work was done while the author was at Ubisoft LaForge.}\;\;
                                Rafael M. O. Cruz$^1$\;\;
        }
        \\
        \\
{\parbox{\textwidth}{\centering  Ecole de Technologie Supérieure$^1$\;\;\; Ubisoft LaForge$^2$\;\;\;
       }
}
\vspace{-20px}
}

\maketitle


\begin{strip}\centering
    \centering
    \includegraphics[width=\linewidth]{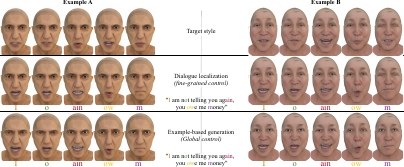}
    \captionof{figure}{KM-speaker enables two types of style control from a source animation (top): (i) dialogue localization (middle) that preserves the source's upper-face expressions while lip-syncing to new audio, and (ii) example-based stylization (bottom) that captures the source's general style. All four generated animations lip-sync to the same target audio but produce distinct results reflecting their respective source styles and control granularities.}
    \label{fig:teaser}
\end{strip}

\begin{abstract}


Speech-driven 3D facial animation methods face significant challenges in simultaneously achieving high-fidelity motion and precise artistic control at production quality. Existing controllable models typically learn global style control by relying on large-scale, low-quality \emph{in-the-wild} datasets that compromise overall animation realism. Furthermore, these frameworks often lack the fine-grained temporal precision required for demanding tasks such as dialogue localization (e.g., dubbing), where matching specific facial expressions is as critical as lip synchronization. We present KM-Speaker (Keypoint-Matching Speaker), a novel keypoint-conditioned flow-based generative framework that provides both global style guidance and frame-level temporal control from reference performances. We propose a disentanglement strategy that separates audio-driven lip motion from keypoint-driven upper-face dynamics, together with a global style context preservation mechanism to ensure coherent full-face expressiveness. KM-Speaker advances example-based 3D facial animation by achieving high-fidelity motion and flexible controllability in a data-constrained setting, consistently outperforming state-of-the-art methods in lip-sync accuracy, style adherence, and expressive temporal control. 

\end{abstract}
\section{Introduction}\label{sec:introduction}

Facial animation plays a crucial role in creating immersive experiences in games and animated films. Traditional production pipelines heavily rely on high-end 4D capture systems, professional actors, and skilled animators to achieve high-fidelity animations \cite{debevec2012light}. This enables directors and designers to follow a precise creative vision by leveraging the organic nuances of real-world speech and expressions. Despite these advantages, this approach is costly, labor-intensive, and inherently rigid, limiting post-capture flexibility and making it impractical for animating large casts of characters or producing content in multiple languages (e.g., dubbing).

Speech‑driven 3D facial animation has emerged as a scalable alternative to traditional capture‑based pipelines, with approaches broadly falling into two categories. Artist‑centric systems (e.g., viseme‑ or curve‑based rigs) \cite{edwards2016jali, zhou2018visemenet, pan2022vocal} offer interpretability and fine control but require extensive manual tuning to achieve high‑quality expressive motion. In contrast, data‑driven methods \cite{xing2023codetalker,xu2024kmtalk,chu2025artalk} learn facial dynamics directly from real performances, offering a significantly higher ceiling for realism. These methods typically operate in vertex space rather than rig‑specific parameter spaces, which limits direct artistic controllability but provides a practical advantage: the generated motion is not tied to a rig parameterization, facilitating integration into different production setups.

A key challenge, therefore, lies in bridging the gap between realism and controllability. This motivates recent work to explore controllable generative frameworks, often conditioning on discrete labels concept tags such as emotions \cite{EMOTE,liu2024emoface,wu2024probtalk3d,liu2025medtalk}. While effective for coarse adjustments, this paradigm is inherently rigid: models are trained on a fixed vocabulary of styles, requiring retraining to introduce new expressions. Moreover, expanding the label set significantly increases annotation cost, encouraging the use of broad categories (e.g., “angry”) that collapse expressive variability into caricature-like behaviors. Alternative approaches introduce control via text or image prompts \cite{zhong2024expclip, zhao2024media2face,wu2024mmhead,chen2025cafe,liu2025medtalk}. However, such modalities struggle to capture subtle behavioral traits, speaker-specific dynamics, and temporal nuances of facial motion, due to language ambiguity and the static nature of images. As a result, achieving precise, expressive, and production-oriented controls over speech-driven facial animation remains an open challenge. 

\begin{figure*}[ht]
    \centering
    \includegraphics[width=\linewidth]{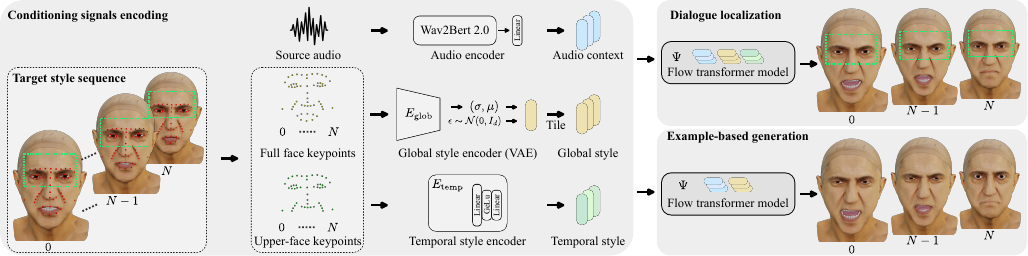}
    \caption{KM‑Speaker architecture and applications. A source audio signal and two sets of target keypoints are processed independently. Full‑face keypoints provide global style features, while upper‑face keypoints provide temporal style cues. Conditioning the flow model $\Psi$ on all inputs enables dialogue localization, where the target upper‑face motion (green boxes) is matched to a new audio clip. Conditioning the same model only on audio and global style enables example‑based generation, where the generation's overall style matches the target.}
    \label{fig:global_architecture_figure}
\end{figure*}

Example-based strategies \cite{sun2024diffposetalk,fu2024mimic,chen2025disenemo,pan2025modelMSMD} offer a compelling alternative by allowing reference videos or animations to guide the global delivery style, greatly simplifying the specification of the desired intent. However, these methods face two key limitations.
First, they rely on large-scale datasets derived from 3D reconstructions \cite{DECA:Siggraph2021, wood20223d} of in-the-wild monocular videos, which significantly compromises quality in two ways: the source videos themselves are often of poor quality (e.g., low resolution, challenging lighting conditions, occlusions), and the parametric morphable models \cite{FLAME:SiggraphAsia2017} used for reconstruction have limited expressivity. This combination results in imprecise facial motion and poor lip sync. While high-end 4D capture systems would address these issues, their cost makes large-scale collection impractical.
Second, they lack the fine‑grained control mechanisms necessary for applications such as dubbing (used interchangeably with dialogue localization throughout this work), where lip-synchronization must be adapted to new audio while preserving the original eyebrow raises, eye closures, and general expressions \cite{bigioi2023multilingual}. 

MeshTalk \cite{richard2021meshtalk} partially addresses dubbing by preserving upper-face motion while generating new lip motion from audio, but it lacks a global style mechanism for ensuring coherent integration of upper- and lower-face motion. In summary, existing methods face a fundamental trade-off between animation quality and controllability, failing to meet production requirements where artists need both high-fidelity and flexible control.

In this paper, we present KM-Speaker (Keypoint Matching Speaker), a flow-based framework \cite{liuflow} for speech-driven facial animation providing both example-based global control and fine-grained temporal controllability  (as illustrated in \cref{fig:teaser}). Unlike prior example-based approaches, our model operates in a realistic data-constrained setting using a curated, high-quality expressive dataset, reflecting practical  data acquisition and usage constraints, while producing accurate and synchronized facial motion. We condition generation on sparse facial keypoints from a reference animation, encoding them as global and temporal style signals, and design a disentanglement strategy to separate lower-face motion from audio and upper-face motion from temporal style. The explicit temporal signal, together with this disentanglement strategy, enables dubbing applications. We also propose a style context preservation mechanism based on modality dropout that enforces reliance on global style to maintain full-face coherence. This enables example-based generation, where the animation adheres to the target's overall expressive intent. 

\noindent The contributions of our paper are as follows:

\begin{itemize}

    \item We introduce KM-Speaker, a keypoint-conditioned framework for speech-driven 3D facial animation. Our model achieves high-fidelity generation with controllable style in data-constrained settings using high-quality expressive data. 
    
    \item Our framework enables both example-based global style control and fine-grained frame-level temporal control from reference performances, supporting production workflows such as dialogue localization.
    
    \item We propose a disentanglement strategy that separates audio-driven lip motion from keypoint-driven upper-face dynamics, combined with a global style context preservation mechanism that maintains coherent full-face expressiveness.
    
    \item Extensive evaluations and user studies demonstrate that our method outperforms state-of-the-art approaches in lip-sync accuracy, style adherence, and dialogue localization.

\end{itemize}

\section{Related work}\label{sec:related_work}


\noindent \textbf{Example-based facial animation generation.} 
Recent literature explores the benefits of example-based generation, aiming to capture the global delivery style of a short target animation sequence (e.g., speaker‑specific idiosyncrasies or characteristic speaking patterns) \cite{fu2024mimic, sun2024diffposetalk, chen2025disenemo, pan2025modelMSMD}. DiffPoseTalk \cite{sun2024diffposetalk} extracts style through a temporally agnostic full‑face encoder–decoder trained with contrastive learning. MIMIC \cite{fu2024mimic} also derives global style from full‑face motion but uses an end‑to‑end approach with a speaker‑classification regularizer, effectively constraining the style space to speaker‑dependent clusters. DisenEmo \cite{chen2025disenemo} further restricts the latent style representation through emotion‑aware supervision. MSMD \cite{pan2025modelMSMD} infers style from implicit latent expression codes from SEREP \cite{josi2024serep} instead, using a VAE to structure the style manifold while relaxing constraints to preserve both inter‑speaker and inter‑emotion variability. Despite their strengths, current example\allowbreak-based methods \cite{fu2024mimic, sun2024diffposetalk, chen2025disenemo, pan2025modelMSMD} rely on large-scale, reconstruction‑based datasets, whose limited fidelity inherently constrains the realism of generated motion. In contrast, our method operates in a challenging yet high-quality data setting. Furthermore, we introduce a temporal style signal, enabling additional controls through dialogue localization, a capability not addressed in prior example-based work.

\noindent \textbf{Dialogue localization.} 
Dialogue localization adapts spoken content in films and video games across languages and cultures, primarily through dubbing. The goal is to preserve the original actor’s performance, such as eye blinks, eyebrow motion, and expressions, while seamlessly adapting lip motion to a new audio \cite{bigioi2023multilingual}. Often referred to as video dubbing in 2D, existing methods fall into either end‑to‑end pipelines \cite{zhang2024musetalk, liu2025identity}, which map source frames and audio directly to the output, or structural approaches \cite{bigioi2022pose} that rely on interpretable intermediate representations, such as facial keypoints. Although structural methods require extra preprocessing (e.g., detection and landmark extraction), they offer improved interpretability and higher‑fidelity expression synthesis \cite{bigioi2023multilingual}. 

Trained on a large-scale capture dataset of 250 individuals, Mesh\allowbreak-Talk \cite{richard2021meshtalk} is the first speech‑driven 3D facial animation framework explicitly designed for dubbing. It conditions a decoder on audio features fused with full‑face temporal embeddings and employs a disentanglement mechanism to separate audio‑driven motion from the target upper‑face dynamics. However, MeshTalk does not enforce full facial expressive coherence, which is essential when the audio alone cannot convey the intended emotional delivery. To address this limitation, we introduce a global style signal that ensures coherent full‑face expression, complementing both audio and target temporal cues. Furthermore, inspired by the demonstrated benefits of keypoints in video dubbing and their effectiveness in 3D settings \cite{wood2022dense, nocentini2023learningS2LS2D}, we introduce a lightweight and explicit style representation based on sparse, semantically meaningful 3D facial landmarks, well‑suited for training in our realistic data constrained setting.

\begin{figure}
    \centering
    \includegraphics[width=1\linewidth]{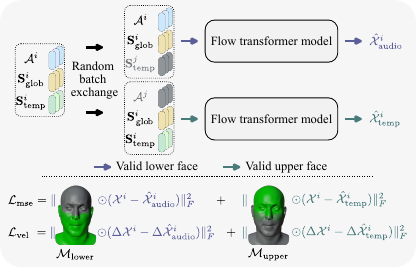}
    \caption{Disentanglement strategy. We randomly exchange either audio $\mathcal{A}^i$ or temporal style $\mathbf{S}_\text{temp}^i$ signals using the corresponding signal from another sample $j$ of the same batch. Coherent lower-face motion is enforced through losses masked by $\mathcal{M}_\text{lower}$, while upper-face coherence is encouraged via $\mathcal{M}_\text{upper}$. Never swapping the global style signal enforces global facial coherence. Masks are visualized on a neutral face, where greener vertices correspond to mask weights closer to 1.    
    }
    \label{fig:disentanglement_strategy} 
\end{figure}

\section{Method}
Our goal is to generate speech-driven facial animation that flexibly matches a reference style. Given input audio and keypoints from a reference animation, we extract two style signals: a time-invariant global style from full-face keypoints captures general style, and a frame-wise temporal style from upper-face keypoints that enables precise upper-face matching. This allows two generation modes: dialogue localization, where all signals condition the model to re-synchronize lip motion while preserving upper-face dynamics, and example-based generation, where only audio and global style condition the model to synthesize full motion matching the reference's overall style. We first provide a general description of our model, including its components and formulation within the flow-matching paradigm (\cref{sec:model_description}). We follow with our audio–keypoint disentanglement strategy  (\cref{sec:disentanglement}) and finally introduce our global style context preservation approach (\cref{sec:context_pres}). 
%
\vspace{-1em}
\subsection{Model architecture}\label{sec:model_description}
The overall architecture of KM-Speaker is illustrated in \cref{fig:global_architecture_figure}. Our flow-based transformer model generates facial motion sequences for fixed window of length $N$ frames conditioned on three features signals: audio context $\mathcal{A}$, global style $\mathcal{S}_\text{glob}$, and temporal style $\mathcal{S}_\text{temp}$. We describe how each feature vector is obtained below.

\noindent \textbf{Audio encoder.} Input speech is encoded using a pretrained audio encoder. We average its hidden-layer features, project them to our model's latent dimension $d$, and resample to match the animation frame rate. The processed audio features are denoted $\mathcal{A} \in \mathbb{R}^{N \times d}$.

\noindent \textbf{Global style encoder.} The global style encoder $E_\text{glob}$ takes a sequence of full-face 3D keypoints $\mathcal{K} \in \mathbb{R}^{N \times 3K}$, where $K$ is the number of keypoints per frame. The encoder first encodes the keypoint signal $\mathcal{K}$ to a sequence of framewise feature embeddings using convolutional layers followed by a transformer encoder, and then applies a temporal average pooling to form a compact global style representation $\mathbf{z}_\text{g}\in \mathbb{R}^{d}$ which captures the reference animation's overall style. This vector is then replicated across all frames to form the global style representation $\mathcal{S}_{glob} = [\mathbf{z}_g]_{n=1}^N \in \mathbb{R}^{N \times d}$. We implement ${E}_\text{glob}$ as a VAE-based encoder following \cite{pan2025modelMSMD}, which predicts a mean and variance regularized via KL divergence.


\noindent \textbf{Temporal style encoder.} The temporal style encoder $E_{\text{temp}}$ encodes frame-wise upper-face dynamics from an upper-face keypoint sequence $\mathcal{K}_\text{u} \in \mathbb{R}^{N \times 3K_\text{u}}$, where $K_\text{u}$ is the number of upper-face keypoints. The encoder produces a temporal style sequence $\mathcal{S}_\text{temp}  \in \mathbb{R}^{N \times d}$.
This signal enables precise frame-level matching of upper-face motion, essential for dialogue localization where lip motion is re-animated while preserving eyebrow raises, gaze shifts, and other upper-face expressions. We implement $E_\text{temp}$ as a two-layer MLP that projects each keypoint frame to the model latent dimension $d$.

\noindent \textbf{Flow transformer model.}
We represent the target facial animation using PCA-compressed vertex offsets. Specifically, we project the per-frame mesh deformations into a low-dimensional latent space of dimension $l$, resulting in a target latent sequence $\mathbf{x} \in \mathbb{R}^{N \times l}$ of latent codes. 
We employ a transformer-based flow model $\mathbf{\Psi}$ parameterizes a continuous transformation from a standard normal prior $p_0$ to plausible facial motion trajectories, inspired by conditional flow matching \cite{lipman2024flow}. This transformation is governed by the ODE $\frac{d\mathbf{x}_t}{dt} = u_t(\mathbf{x}_t)$, where $\mathbf{x}_t \in \mathbb{R}^{N \times l}$ denotes the latent sequence at flow time $t \in [0,1]$. We construct trajectories using conditional optimal transport paths~\cite{lipman2023flow}, with linear interpolation $\mathbf{x}_t = (1-t)\mathbf{x}_0 + t\mathbf{x}_1$, where $\mathbf{x}_0 \sim \mathcal{N}(0, \mathbf{I})$. Rather than directly regressing velocities, we adopt an $\mathbf{x}_1$-prediction parameterization: $\mathbf{\Psi}(\mathbf{x}_t, t, \mathbf{c})$ predicts the clean target $\mathbf{x}_1$ from $\mathbf{x}_t$, conditioned on $\mathbf{c} = (\mathcal{A}, \mathbf{S}_\text{glob}, \mathbf{S}_\text{temp})$. This enables supervision through vertex-based losses without ODE integration during training. At inference, the predicted $\hat{\mathbf{x}}_1$ is converted to a velocity estimate $\hat{u}_t = (\hat{\mathbf{x}}_1 - \mathbf{x}_t)/(1-t)$, and the ODE is solved (midpoint, 50 steps) to produce latent coefficients $\hat{\mathbf{x}} \in \mathbb{R}^{N \times l}$. Latents are decoded to vertex space via the PCA basis $\mathcal{R}_{\text{PCA}}: \mathbb{R}^{N \times l} \rightarrow \mathbb{R}^{N \times 3V}$, where $V$ is the number of mesh vertices, yielding the final animation $\hat{\mathcal{X}} = \mathcal{R}_{\text{PCA}}(\hat{\mathbf{x}})$.


\subsection{Audio-keypoint disentanglement}\label{sec:disentanglement}

We propose a disentanglement strategy to achieve three objectives: \textbf{(1)} audio must control lip synchronization and mouth shapes for natural delivery, \textbf{(2)} temporal style must precisely drive upper-face motion frame-by-frame according to the target animation, enabling dialogue localization, and \textbf{(3)} global style must govern overall delivery intent and enforce coherent full-face motion.

Inspired by MeshTalk \cite{richard2021meshtalk}, we define per-vertex weight masks $\mathcal{M}_{\text{upper}}, \mathcal{M}_{\text{lower}}$ with weights in $[0,1]$ that partition the face into upper and lower regions. We extend their approach by incorporating style conditioning to ensure facial coherence (\cref{fig:disentanglement_strategy}).

During training, for each sample $i$ in a batch of size $B$, we define a conditioning tuple $\mathbf{c}^i = (\mathcal{A}^i, \mathcal{S}_{\text{glob}}^i, \mathcal{S}_{\text{temp}}^i)$ corresponding to target animation $\mathcal{X}^i$. To enforce disentanglement, we randomly select another sample $j \neq i$ from the batch and create two mismatched conditioning variants: 1) \textbf{Audio conditioning:} $\mathbf{c}^i_{\text{audio}} = (\mathcal{A}^i, \mathcal{S}_{\text{glob}}^i, \mathcal{S}_{\text{temp}}^j)$: keeps audio from $i$, swaps temporal style from $j$. 2) \textbf{Temporal conditioning:} $\mathbf{c}^i_{\text{temp}} = (\mathcal{A}^j, \mathcal{S}_{\text{glob}}^i, \mathcal{S}_{\text{temp}}^i)$: swaps audio from $j$, keeps temporal style from $i$. Crucially, the global style $\mathcal{S}_{\text{glob}}^i$ always matches the target animation $\mathcal{X}^i$, ensuring coherent integration of upper- and lower-face motion even when audio and temporal style intent is different (objective \textbf{(3)}).

We generate predictions for both conditioning variants: $\hat{\mathcal{X}}_\text{audio}^i=\mathcal{R}_\text{PCA}(\Psi(\mathbf{x}_t^i, t, \mathbf{c}^i_\text{audio}))$ enforces to match $\mathcal{X}^i$ in the \emph{lower face} (audio drives lips, objective \textbf{(1)}) and $\hat{\mathcal{X}}_\text{temp}^i=\mathcal{R}_\text{PCA}(\Psi(\mathbf{x}_t^i, t, \mathbf{c}^i_\text{temp}))$, which enforces to match $\mathcal{X}^i$ in the \emph{upper face} (temporal style drives upper face, objective \textbf{(2)}).

\noindent\textbf{Training losses.} We train the model using a disentangled reconstruction loss:

\begin{equation}
\begin{aligned}
\mathcal{L}_{\text{mse}} =
&\lVert\mathcal{M}_{\text{lower}} \odot (\mathcal{X}^i - \hat{\mathcal{X}}_{\text{audio}}^i)\rVert^2_F \\
&+ \lVert \mathcal{M}_{\text{upper}} \odot (\mathcal{X}^i - \hat{\mathcal{X}}_{\text{temp}}^i) \rVert^2_F
\end{aligned}
\end{equation}

\noindent where $\odot$ denotes the Hadamard product and $\lVert \cdot \rVert_F$ is the Frobenius norm.

To regularize temporal smoothness, we add a velocity loss \cite{Cudeiro_2019_CVPRVOCA} adapted for our disentanglement strategy:


\begin{equation}
\begin{aligned}
\mathcal{L}_{\text{vel}} =
& \mathcal{M}_{\text{lower}} \odot (\Delta\mathcal{X}^i - \Delta\hat{\mathcal{X}}_{\text{audio}}^i) \rVert^2_F \\
& + \lVert \mathcal{M}_{\text{upper}} \odot (\Delta\mathcal{X}^i - \Delta\hat{\mathcal{X}}_{\text{temp}}^i) \rVert^2_F
\end{aligned}
\end{equation}

\noindent where $\Delta\mathcal{X} = \mathcal{X}_{1:N} - \mathcal{X}_{0:N-1}$.

Finally, we regularize the global style encoder's latent space with KL divergence: $\mathcal{L}_{\text{KL}} = \frac{1}{2}\sum_{i=1}^{d}\left(\mu_i^2 + \sigma_i^2 - \log(\sigma_i^2) - 1\right).$

The total training loss is:
\begin{equation}
    \mathcal{L} = \mathcal{L}_{\text{mse}} + \lambda_{\text{vel}} \mathcal{L}_{\text{vel}} + \lambda_{\text{KL}} \mathcal{L}_{\text{KL}},
\end{equation}
where $\lambda_{\text{vel}}$ and $\lambda_{\text{KL}}$ are hyperparameters balancing each loss term.

\subsection{Global style preservation}\label{sec:context_pres}
The disentanglement strategy described above creates an asymmetry in conditioning strength: the temporal style signal $\mathcal{S}_{\text{temp}}$ provides rich, frame-wise upper-face motion information that can dominate the generation, while the global style $\mathcal{S}_{\text{glob}}$ encodes only high-level information. This imbalance poses a problem for example-based generation. At inference, when temporal style is absent (i.e., when we want to match the reference's style), the model may fail to leverage $\mathcal{S}_{\text{glob}}$ effectively for upper-face generation as it has learned to rely primarily on the stronger temporal signal during training.

To address this, we randomly zero out the temporal style signal while always retaining the global style. This forces the model to learn to capture the overall style from $\mathcal{S}_{\text{glob}}$ alone when temporal information is unavailable, ensuring that global style remains influential even in the presence of strong temporal conditioning. Combined with our disentanglement strategy, this dropout naturally creates training samples with conditioning tuples $(\mathcal{A}^j, \mathcal{S}_{\text{glob}}^i, \varnothing)$ paired with target $\mathcal{X}^i$. This configuration encourages the model to maintain the target's global style across arbitrary audio inputs, including those with conflicting emotions.
\section{Experimental protocol}\label{sec:experimental_protocol}
Our experimental protocol is designed to assess lip‑motion accuracy, global style transfer fidelity, and the model’s dubbing capabilities. \\


\noindent \textbf{Dataset.} 
Our dataset consists of 2.6 hours of 4D captured data in consistent topology, evenly distributed across 12 professional actors and recorded at 60 FPS. Each actor performs a personalized set of sentences spanning 8 emotional states and 2 intensities (see capture protocol details in Appendix \ref{sup:data_collection_protocol}). We do not leverage label information during training to preserve diversity within each emotional state (\cref{sec:introduction}). We use data from 8 actors for training, 1 for validation, and 3 for testing. 


\noindent \textbf{Baseline models.} For global stylization evaluation, we compare our method to the state‑of‑the‑art example‑based models MIMIC \cite{fu2024mimic} (deterministic) and MSMD \cite{pan2025modelMSMD} (diffusion\allowbreak‑based). Both methods are originally learned on large-scale datasets, encode style from full-face, and accept an audio and a target window animation for style extraction. We compare against their original released models, as well as their retrained and finetuned versions on our small high-quality dataset (respectively noted \_retrain or \_finetune) for fair comparison. These variants assess the impact of data sources on style control and lip synchronization. For dialogue localization evaluation, we compare against the original MeshTalk \cite{richard2021meshtalk} model, which to our knowledge is the only prior work explicitly addressing this task. We also evaluate the impact of the style input through training with a dense set of 1660 keypoints derived from \cite{wood2022dense} (denoted Ours\_1660\_keypoints) and a sparse set of 68 keypoints commonly used for facial tasks \cite{sagonas2016300} (noted Ours\_68\_keypoints) visually presented \cref{fig:keypoints_and_masks}. In all experiments, bold entries indicate statistically significant improvements (Wilcoxon signed rank test, $\rho<0.01$).

\noindent \textbf{Implementation details.} 
Our audio encoder relies on w2v-BERT 2.0 \cite{barrault2023seamlessw2vbert2}. 
Our global style encoder is formed of 1D convolutions followed by transformer layers, following MIMIC and MSMD style encoder architectures \cite{fu2024mimic, pan2025modelMSMD}. The temporal style encoder is an MLP, while our flow model is a transformer-based model conditioned using style-adaptive layer normalization \cite{min2021meta}. We trained our full model using Adam optimizer \cite{kingma2014adam} for 2.3 days on a single RTX 8000 GPU. More details on the training, architecture, and mask construction are provided in Appendix \ref{sup:training_details} (training), \ref{sup:model_and_representations} (architecture), and \ref{sup:facial_region_masks} (masks).

We preserved the original MIMIC and MSMD protocols (25 FPS training) when training and finetuning those models on our dataset. MSMD and MIMIC use 100-frame style context windows at 25 FPS, whereas our method uses 200-frame context windows at 60 FPS. MIMIC\allowbreak\_finetune is trained in the FLAME topology \cite{FLAME:SiggraphAsia2017} and the output animation is converted back to our topology for evaluation, whereas MIMIC\allowbreak\_retrain is directly trained in our own topology. For the MeshTalk model, we convert the input neutral mesh and target style animations to their topology prior to generation, and revert the topology to our own for evaluation. Topology conversion details are provided in Appendix \ref{sup:topology_conversion}. For fair comparison, example-based experiments are conducted at 25 FPS, and dialogue localization experiments at 30 FPS. 

Unless stated otherwise, rendered animations are augmented with textures, eyes, and teeth to improve perceptual realism. Details on assets are provided in Appendix \ref{sup:auxiliary_components_for_rendering}. We provide additional geometry-only visualizations in Appendix \ref{sup:raw_geometry_results}.

\begin{table}[t]
\centering 
\caption{Quantitative evaluation on our held-out set of actors in a matching audio and style context. We present baseline models' performance and ablations regarding the signal passed to our global style encoder.}
\scalebox{0.88}{
\begin{tabular}{lcccc}
\textbf{Model} & \textbf{LVE} & \textbf{MSE} & \textbf{MOD} & \textbf{FDD} \\
               & {\small (mm)$\downarrow$} & {\small (mm)$\downarrow$} & {\small (mm)$\downarrow$} & {\small (mm$^{-2}$)$\downarrow$} \\
\midrule
MIMIC \cite{fu2024mimic}     & 2.48 & 2.00 & 0.68 & 4.68 \\
MIMIC\_finetune                    & 2.41 & 1.97 & \textbf{0.47} & 4.59 \\
MIMIC\_retrain                    & 0.85 & 0.49 & 0.50 & 4.69 \\
MSMD \cite{pan2025modelMSMD} & 1.96 & 1.10  & 0.91 & 9.38 \\
MSMD\_finetune                     & 1.35 & 0.73 & 0.62 & 6.38 \\
MSMD\_retrain                     & 1.11 & 0.65 & 0.56 & 4.51 \\

\midrule

ours\_meshtalk\_encoder                 & 1.10 & 0.65 & 0.50 & 5.63 \\
ours\_1660\_keypoints                 & 0.83 & 0.49 & \textbf{0.46} & 5.23 \\
ours\_68\_keypoints                & \textbf{0.78} & 0.47 & \textbf{0.45} & 4.52 \\




\midrule
ours                         & \textbf{0.77} & \textbf{0.45} & \textbf{0.46} & \textbf{4.31} \\
\bottomrule
\end{tabular}}
\label{tab:previous_window_quantitative}
\end{table}

\section{Results}


\subsection{Matching audio-style context}\label{sec:matching_audio_style}
We first quantitatively evaluate example-based models under matching audio and style contexts (e.g., same emotion), measuring lip-sync and motion accuracy. Using all sequences from our test set, we condition on the previous window as global style and the current window as audio, with no overlap to prevent leakage \cite{pan2025modelMSMD}. Note that no temporal style is used for example-based generations.


We evaluate generation accuracy using standard metrics: vertex MSE (global motion), LVE \cite{richard2021meshtalk} (lip accuracy), MOD \cite{sun2024diffposetalk} (mouth aperture), and FDD \cite{xing2023codetalker} (upper-face dynamics). Results for all baselines and related ablations are reported in \cref{tab:previous_window_quantitative}.


Baseline results for the MIMIC and MSMD variants highlight the benefits of finetuning and training on higher‑quality data, yielding consistent improvements across both lip‑related metrics (LVE, MOD) and global motion (MSE) compared to the author‑released versions. Upper‑face dynamics (FDD) also improve for MSMD, whereas they remain comparatively stable for MIMIC, which we attribute to MIMIC's deterministic nature that constrains stochastic events such as blinks and brow motion.

Overall, our model achieves significant improvements over baseline methods thanks to the efficiency of our flow‑based formulation and the use of a sparse yet explicit global‑style keypoint signal. This is supported by ablation results (\cref{tab:previous_window_quantitative}) showing that substituting our keypoint representation with implicit embeddings or denser keypoint configurations reduces general performance. For instance, using implicit representations derived from the MeshTalk encoder as input to our global style encoder yields the lowest scores across all ablation metrics. Then, reducing the keypoint density from 1660 to 68 and our 55 keypoints (as detailed \Cref{sec:experimental_protocol}) while preserving semantically meaningful facial locations or refining them by introducing nasolabial folds keypoints produces a more explicit and informative style cue, consistently benefiting the model. This likely facilitates encoding, yielding a compact representation better suited to our challenging data setting. Notably, MSMD derives its style features from implicit expression codes computed via SEREP \cite{josi2024serep}, while MIMIC operates on full‑face inputs, both of which are conceptually aligned with the implicit and dense alternatives tested in our ablations. 

\begin{figure}
    \centering
    \includegraphics[width=\linewidth]{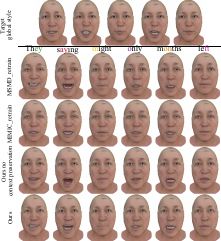}
    \caption{Generation for the different baselines with a desired angry target style sequence while the audio intent is mostly neutral. }
    \label{fig:example_based_figure}
\end{figure}

\subsection{Cross audio-style context}\label{sec:global_style_robustness}
We qualitatively evaluate our model against baselines trained on the our high‑quality dataset (MSMD\allowbreak\_retrain and MIMIC\allowbreak\_retrain) as well as an ablated version of our method without the context‑preservation mechanism (\cref{sec:context_pres}). The goal is to assess how well each model adheres to a target animation’s global style when the audio does not match the target intent. Such a scenario is common in production, where style nuances (e.g., suspicion, doubt, or irony conveyed through eyebrow) are absent from the audio. \Cref{fig:example_based_figure} illustrates this by using a desired angry target style paired with a neutral audio line. We provide additional qualitative examples in supplementary video (3:33 to 4:31).

By balancing the importance of conditioning signals and training with conflicting audio–style pairs (\cref{sec:context_pres}), our model produces upper‑face motion that more faithfully reflects the target style. This improvement is especially visible when comparing our model's expressive upper-face intent to its less expressive ablated variant. Similarly, MIMIC\allowbreak\_retrain and MSMD\allowbreak\_retrain remain mostly neutral, which we believe stems from the challenging size of our dataset that hinders their ability to capture style.

\subsection{User study}

We conducted two user studies to evaluate (1) lip synchronization quality and (2) global style adherence across three models: Ours, MIMIC\allowbreak\_retrain, and MSMD\allowbreak\_retrain. We include textures, as well as eye and teeth assets for more natural assessment and ensure fair comparison by comparing models trained on our dataset (best-performing variants in our quantitative evaluation \cref{sec:matching_audio_style}). 

\begin{figure}[!t]
    \centering
    \includegraphics[width=\linewidth]{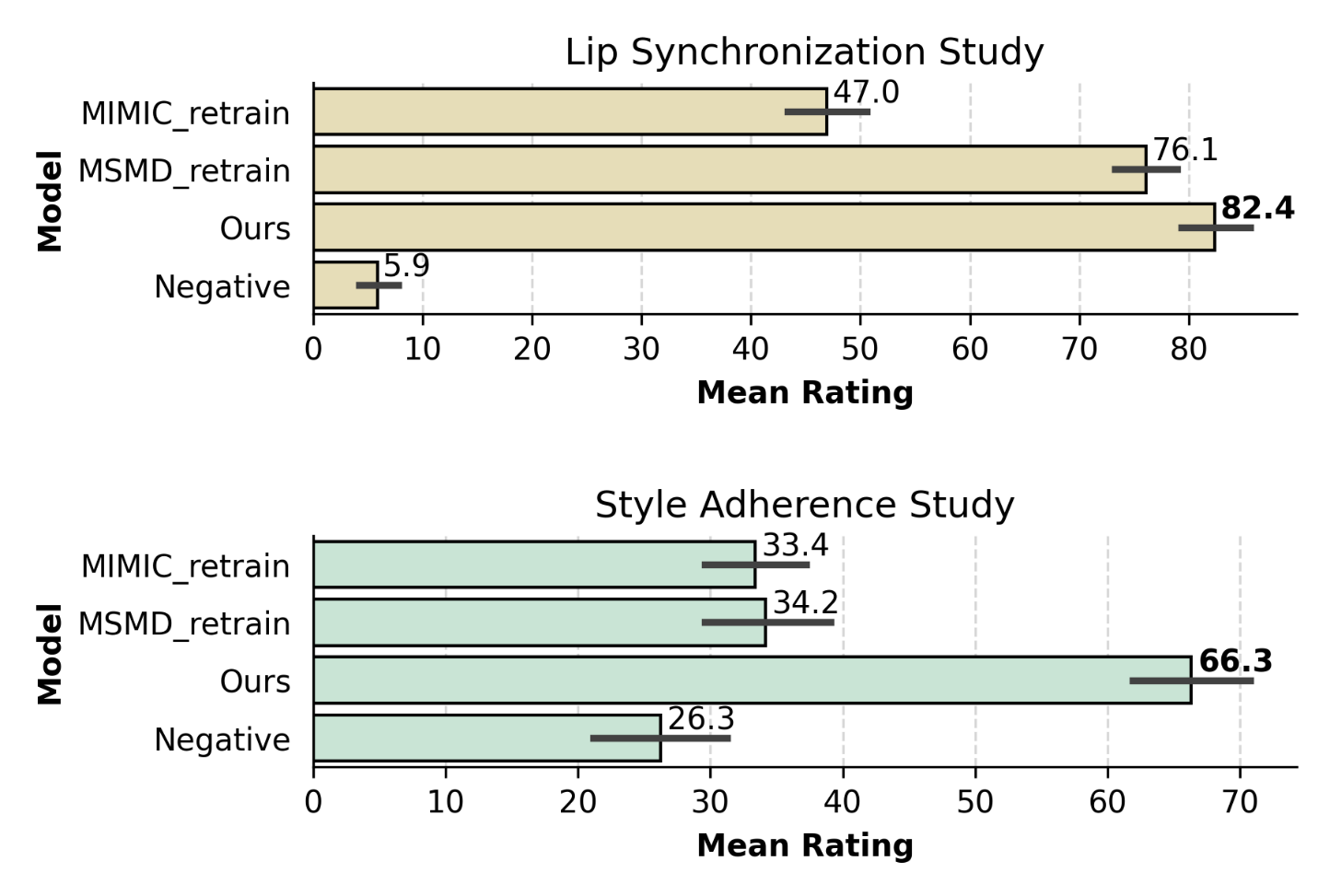}
    \caption{Lip‑synchronization (25 responses) and style‑adherence user study (23 responses) results. We report mean ratings ($95\%$ CI). }
    \label{fig:user_study_plots}
    \vspace{-1em}
\end{figure}

Each study comprises five sets of ten randomly selected, non-overlapping videos covering all test actors. Samples from the lip-sync and style-adherence studies are drawn from pools of 83 lip-sync and 200 style examples, respectively, with randomized model ordering. We adopt a MUSHRA-like protocol \cite{series2014method}, where participants rate samples on a 0–100 scale, and a negative anchor is hidden in each set. Lip-sync evaluation includes audio, whereas style evaluation is muted, following \cite{pan2025modelMSMD}, to improve focus on style consistency. Studies were conducted with researchers from the video‑game industry and professionals in animation, yielding 25 and 23 responses for the lip-synchronization and style-adherence tasks, respectively. We report the mean rating ($95\%$ CI) in \cref{fig:user_study_plots}. Study guidelines, structure, visuals, anchor selection, and additional analyses (i.e., mean rankings and agreement measures) are provided in Appendix \ref{sup:perceptual_study}.


Results show our method as preferred for lip-synchronization, with MIMIC\allowbreak\_retrain lagging significantly behind, and MSMD\allowbreak\_retrain scoring closely behind. This ranking differs from the quantitative close-context scenario, where lip-related metrics (LVE) show MIMIC\allowbreak\_retrain performing better than MSMD\allowbreak\_retrain. We attribute this inversion to the limitations of LVE, which measures strict geometric distance to a single ground-truth reference. Because speech-to-motion is a one-to-many mapping, generative models can produce plausible, high-quality lip articulations that diverge from the specific actor’s performance, yet are penalized by this metric. Visually, MIMIC\allowbreak\_retrain presents a less organic, expressive, and more damped motion than MSMD, further supported by qualitative results in \cref{fig:previous_window_same_context}. While this conservative behavior benefits MIMIC\allowbreak\_retrain regarding LVE (by remaining closer to the mean), it degrades perceptual quality. In comparison, our model achieves better lip-synchronization than all competitors in both the user study and quantitative measures, confirming the benefits of the flow-based approach in delivering both geometrically accurate and perceptually convincing results.

Across the style-adherence evaluation, MIMIC\allowbreak\_retrain  and MSMD\allowbreak\_retrain underperform markedly, consistent with our qualitative observations in \cref{sec:global_style_robustness} that these models struggle to capture style when trained on small datasets (see additional examples in supplementary video 3:33). Upon inspection, we observe that MIMIC\allowbreak\_retrain and MSMD\allowbreak\_retrain frequently produce either near-neutral or aligned with the speech emotion deliveries, both of which deviate from the target style and are therefore similarly penalized.

\subsection{Dialogue localization}\label{sec:dialogue_localization}

\begin{figure*}
    \centering
    \includegraphics[width=1\linewidth]{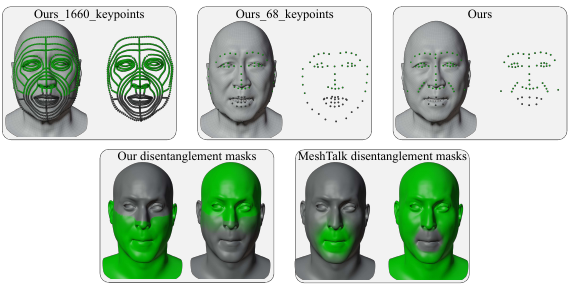}
    \caption{We visually present the keypoints used in Ours\_1660\_keypoints, Ours\_68\_keypoints, and Ours (top row), along with the disentanglement masks from our method and from MeshTalk \cite{richard2021meshtalk} (bottom row). Green keypoints denote the upper‑face keypoints used by the temporal encoder model, while the combination of green and grey keypoints represents the full‑face keypoint signal used as style input. Disentanglement masks correspond to per‑vertex weights, where grey vertices indicate weights of zero and green vertices indicate non‑zero weights, with more intense green representing weights closer to 1.}
    \label{fig:keypoints_and_masks}
\end{figure*}

\begin{figure*}
        \centering
        \includegraphics[width=1\linewidth]{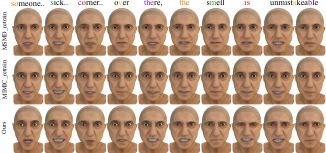}
        \caption{Qualitative comparison of MSMD\_retrain, MIMIC\_retrain, and Ours in the matching context scenario, where the target style matches the audio intent. We do not display the target to encourage focus on the lip-sync and naturalness.}
        \label{fig:previous_window_same_context}
\end{figure*}

We evaluate dialogue localization: matching target expressions while resynchronizing lip motion to new audio. For quantitative evaluation, we randomly sample 100 pairs of audio and target facial animations of the same length. We report the upper-face mean squared error (U-MSE) measured with respect to the target ground-truth animation and the Lip Vertex Error (LVE) with respect to the corresponding audio ground-truth animation in \cref{tab:dialogue_localization_quantitative}. We compare MeshTalk \cite{richard2021meshtalk} against our model and three ablations: using the MeshTalk full-face encoder instead of keypoints, using MeshTalk disentanglement masks (that allow the target animation to affect jaw motion \cref{fig:keypoints_and_masks}), and our model trained with no disentanglement strategy. 

Our model achieves the lowest LVE while best preserving target upper-face motion, indicating superior dialogue localization performance. The temporal keypoint signal provides strong supervision: without disentanglement (Ours\allowbreak\_no\allowbreak\_disentangle), it leaks into lip motion, degrading lip-sync quality. MeshTalk’s disentanglement partially mitigates this effect but still allows upper-face dynamics to influence lip articulation via jaw motion, which is detrimental in expressive scenarios. MeshTalk also exhibits lower fidelity to ground-truth motion, likely due to less expressive training data and the absence of mechanisms enforcing coherent full-face expressiveness. Replacing explicit keypoint conditioning with a learned full-face embedding (Ours\allowbreak\_meshtalk\allowbreak\_encoder) also degrades performance, highlighting the benefits of our explicit signal for robust style encoding in constrained data settings.

\begin{figure*}
    \centering
    \includegraphics[width=\linewidth]{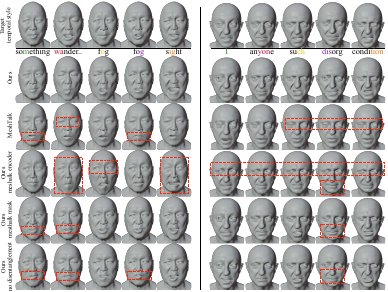}

        \caption{Qualitative results for two different actors and target temporal style for the dialogue localization task. We compare performance between MeshTalk \cite{richard2021meshtalk} and our model, and also conduct ablations on MeshTalk components relative to our approach through: Ours\_meshtalk\_encoder, Ours\_meshtalk\_masks, and Ours\_no\_disentanglement. The red rectangles clearly identify mismatches between the target upper face and the corresponding generated frame, or inaccurate lip shapes with respect to the pronounced sound. Our method precisely matches the target animation's upper face, including challenging eyebrow motion and eye closure.}
    \label{fig:qualitative_resutls_dialogue_localization}
\end{figure*}

We complement these findings with qualitative results in \cref{fig:qualitative_resutls_dialogue_localization}, showing raw geometry outputs without eyes, teeth, or textures. This setup follows the MeshTalk setup and provides a complementary perspective to our previous renderings by isolating predicted geometry motion from rendering assets. The issues described above are clearly visible in dynamic sequences (see supplementary video 5:45–7:07) and, to a lesser extent, in  \cref{fig:qualitative_resutls_dialogue_localization}. Overall, these results highlight the importance of explicit keypoint conditioning and principled disentanglement for coherent localization.

 We provide additional qualitative analysis comparing against a baseline that deterministically blends mouth motion, with or without global style conditioning, with a fixed upper-face animation (see Appendix \ref{sup:blending_baseline}). This setup is inspired by common practices in artist-driven pipelines (e.g., viseme-based systems such as JALI \cite{edwards2016jali}), in which mouth motion is often edited or blended independently of upper-face animation \cite{edwards2016jali}. The comparison highlights inherent limitations of such strategies, which often require time-consuming manual editing and struggle to generalize across diverse expressions and geometries. These observations further support our unified approach, which enables both example-based generation and dialogue localization within a single framework.

\begin{table}[t]
\centering
\caption{Dialogue localization quantitative comparison of the proposed model, MeshTalk, and related ablations.} 
\scalebox{0.90}{
\begin{tabular}{lccc}
\textbf{Models} & \textbf{U-MSE} {\small (mm$^{-1}$)$\downarrow$} & \textbf{LVE} {\small (mm)$\downarrow$} \\
\midrule
MeshTalk \cite{richard2021meshtalk} & 6.60 & 1.68 \\
Ours\_meshtalk\_encoder & 5.64 & 1.42 \\
Ours\_meshtalk\_masks & 2.81 & 1.22 \\
Ours\_no\_disentangle & 2.83 & 1.32 \\
Ours & \textbf{2.77} & \textbf{1.02} \\
\bottomrule
\end{tabular}}
\vspace{-1em}
\label{tab:dialogue_localization_quantitative}
\end{table}

\section{Limitations and future works}

While effective in our high-fidelity setting, training on 8 actors with production-quality audio limits the diversity of the learned facial prior and the range of styles and acoustic conditions, potentially introducing bias toward specific morphologies, expressions, or speech characteristics. This limitation is observed in our generalization experiment (Appendix \ref{sup:generalization} and supplementary video 7:20), where certain facial morphologies lead to imperfect mouth closure or inner-mouth mesh intersections. Addressing these limitations calls for exploring few-shot adaptation to better capture actor-specific attributes. Despite this, our model shows reasonable robustness to in-the-wild styles and audio. Beyond these limitations, our method raises ethical considerations related to example-based generation and dubbing. Such systems should only be used with appropriate consent from the individuals whose likeness or performance is reproduced.

\section{Conclusion}

In this paper, we present KM-Speaker, a flow-based generative framework with the explicit goal of making data-driven models more amenable to artistic direction. To this end, we introduce the first data-driven method enabling both example-based generation and dialogue localization, combining captured realism with flexible control. Leveraging a keypoint-conditioned architecture and a disentanglement strategy, KM-Speaker achieves accurate lip synchronization while capturing global and temporal cues from reference performances. We view this approach as a step toward production-ready solutions that give artists full control over performance while leveraging the realism learned from captured data at scale.


%
\section{Acknowledgements}

This research was supported by the \href{https://www.nserc-crsng.gc.ca/index_eng.asp}{Natural Sciences and Engineering Research Council of Canada (NSERC)} and \href{https://www.mitacs.ca/}{MITACS} through the \textit{Alliance-Mitacs} program (Grant No. ALLRP 589317 - 23).

We would also like to thank the authors of MeshTalk \cite{richard2021meshtalk}, MIMIC \cite{fu2024mimic}, and MSMD \cite{pan2025modelMSMD} for releasing their code and advancing research in the field. 

We also thank the actors who consented to the use of their likenesses for the qualitative results presented in this paper.


{\small
\bibliographystyle{ieee_fullname}
\bibliography{egbib}

@String(CVPR= {IEEE Conf. Comput. Vis. Pattern Recog.})

@String(TOG= {ACM Trans. Graph.})

@String(ICIP = {IEEE Int. Conf. Image Process.})

@String(ICLR = {Int. Conf. Learn. Represent.})

@String(AAAI = {AAAI})

@String(VR   = {Vis. Res.})

@String(CVPR  = {CVPR})

@String(TOG   = {ACM TOG})

@String(ICIP  = {ICIP})

@String(ICLR  = {ICLR})

@inproceedings{EMOTE,
author = {Dan\v{e}\v{c}ek, Radek and Chhatre, Kiran and Tripathi, Shashank and Wen, Yandong and Black, Michael and Bolkart, Timo},
title = {Emotional Speech-Driven Animation with Content-Emotion Disentanglement},
year = {2023},
isbn = {9798400703157},
publisher = {Association for Computing Machinery},
address = {New York, NY, USA},
url = {https://doi.org/10.1145/3610548.3618183},
doi = {10.1145/3610548.3618183},
booktitle = {SIGGRAPH Asia 2023 Conference Papers},
articleno = {41},
numpages = {13},
location = {Sydney, NSW, Australia},
series = {SA '23}
}

@article{FLAME:SiggraphAsia2017, 
  title = {Learning a model of facial shape and expression from {4D} scans}, 
  author = {Li, Tianye and Bolkart, Timo and Black, Michael. J. and Li, Hao and Romero, Javier}, 
  journal = {ACM Transactions on Graphics, (Proc. SIGGRAPH Asia)}, 
  volume = {36}, 
  number = {6}, 
  year = {2017}, 
  pages = {194:1--194:17},
  url = {https://doi.org/10.1145/3130800.3130813} 
}

@article{DECA:Siggraph2021,
author = {Feng, Yao and Feng, Haiwen and Black, Michael J. and Bolkart, Timo},
title = {Learning an animatable detailed 3D face model from in-the-wild images},
year = {2021},
issue_date = {August 2021},
publisher = {Association for Computing Machinery},
address = {New York, NY, USA},
volume = {40},
number = {4},
issn = {0730-0301},
url = {https://doi.org/10.1145/3450626.3459936},
doi = {10.1145/3450626.3459936},
journal = {ACM Trans. Graph.},
month = jul,
articleno = {88},
numpages = {13}
}

@inproceedings{wood20223d,
  title={3d face reconstruction with dense landmarks},
  author={Wood, Erroll and Baltru{\v{s}}aitis, Tadas and Hewitt, Charlie and Johnson, Matthew and Shen, Jingjing and Milosavljevi{\'c}, Nikola and Wilde, Daniel and Garbin, Stephan and Sharp, Toby and Stojiljkovi{\'c}, Ivan and others},
  booktitle={European Conference on Computer Vision},
  pages={160--177},
  year={2022},
  organization={Springer}
}

@article{debevec2012light,
  title={The light stages and their applications to photoreal digital actors},
  author={Debevec, Paul},
  journal={SIGGRAPH Asia},
  volume={2},
  number={4},
  pages={1--6},
  year={2012}
}

@inproceedings{zhao2024media2face,
  title={Media2face: Co-speech facial animation generation with multi-modality guidance},
  author={Zhao, Qingcheng and Long, Pengyu and Zhang, Qixuan and Qin, Dafei and Liang, Han and Zhang, Longwen and Zhang, Yingliang and Yu, Jingyi and Xu, Lan},
  booktitle={ACM SIGGRAPH 2024 conference papers},
  pages={1--13},
  year={2024}
}

@inproceedings{wood2022dense,
  title={3d face reconstruction with dense landmarks},
  author={Wood, Erroll and Baltru{\v{s}}aitis, Tadas and Hewitt, Charlie and Johnson, Matthew and Shen, Jingjing and Milosavljevi{\'c}, Nikola and Wilde, Daniel and Garbin, Stephan and Sharp, Toby and Stojiljkovi{\'c}, Ivan and others},
  booktitle={European Conference on Computer Vision},
  pages={160--177},
  year={2022},
  organization={Springer}
}

@inproceedings{zhu2022celebvhq,
  title={CelebV-HQ: A large-scale video facial attributes dataset},
  author={Zhu, Hao and Wu, Wayne and Zhu, Wentao and Jiang, Liming and Tang, Siwei and Zhang, Li and Liu, Ziwei and Loy, Chen Change},
  booktitle={European conference on computer vision},
  pages={650--667},
  year={2022},
  organization={Springer}
}

@InProceedings{kingma2014adam,
  author    = {Kingma, Diederik and Ba, Jimmy},
  booktitle = {International Conference on Learning Representations (ICLR)},
  title     = {Adam: A Method for Stochastic Optimization},
  year      = {2015},
  address   = {San Diega, CA, USA},
  optmonth  = {12},
}

@misc{triplegangers,
	author = {},
	title = {{T}riplegangers {F}ace {M}odels},
	howpublished = {\url{https://triplegangers.com/}},
	year = {2024},
	note = {Online; Accessed: 13-11-2024},
}

@inproceedings{wu2024probtalk3d,
  title={ProbTalk3D: Non-Deterministic Emotion Controllable Speech-Driven 3D Facial Animation Synthesis Using VQ-VAE},
  author={Wu, Sichun and Haque, Kazi Injamamul and Yumak, Zerrin},
  booktitle={Proceedings of the 17th ACM SIGGRAPH Conference on Motion, Interaction, and Games},
  pages={1--12},
  year={2024}
}

@inproceedings{pan2025modelMSMD,
  title={Model See Model Do: Speech-Driven Facial Animation with Style Control},
  author={Pan, Yifang and Singh, Karan and Hafemann, Luiz Gustavo},
  booktitle={Proceedings of the Special Interest Group on Computer Graphics and Interactive Techniques Conference Conference Papers},
  pages={1--10},
  year={2025}
}

@article{sun2024diffposetalk,
  title={Diffposetalk: Speech-driven stylistic 3d facial animation and head pose generation via diffusion models},
  author={Sun, Zhiyao and Lv, Tian and Ye, Sheng and Lin, Matthieu and Sheng, Jenny and Wen, Yu-Hui and Yu, Minjing and Liu, Yong-jin},
  journal={ACM Transactions on Graphics (TOG)},
  volume={43},
  number={4},
  pages={1--9},
  year={2024},
  publisher={ACM New York, NY, USA}
}

@article{liu2025medtalk,
  title={MEDTalk: Multimodal Controlled 3D Facial Animation with Dynamic Emotions by Disentangled Embedding},
  author={Liu, Chang and Pan, Ye and Ding, Chenyang and Rahardja, Susanto and Yang, Xiaokang},
  journal={arXiv preprint arXiv:2507.06071},
  year={2025}
}

@inproceedings{richard2021meshtalk,
  title={Meshtalk: 3d face animation from speech using cross-modality disentanglement},
  author={Richard, Alexander and Zollh{\"o}fer, Michael and Wen, Yandong and De la Torre, Fernando and Sheikh, Yaser},
  booktitle={Proceedings of the IEEE/CVF international conference on computer vision},
  pages={1173--1182},
  year={2021}
}

@inproceedings{nocentini2023learningS2LS2D,
  title={Learning landmarks motion from speech for speaker-agnostic 3D talking heads generation},
  author={Nocentini, Federico and Ferrari, Claudio and Berretti, Stefano},
  booktitle={International Conference on Image Analysis and Processing},
  pages={340--351},
  year={2023},
  organization={Springer}
}

@inproceedings{wu2024mmhead,
  title={Mmhead: Towards fine-grained multi-modal 3d facial animation},
  author={Wu, Sijing and Li, Yunhao and Yan, Yichao and Duan, Huiyu and Liu, Ziwei and Zhai, Guangtao},
  booktitle={Proceedings of the 32nd ACM International Conference on Multimedia},
  pages={7966--7975},
  year={2024}
}

@inproceedings{zhong2024expclip,
  title={Expclip: Bridging text and facial expressions via semantic alignment},
  author={Zhong, Yicheng and Wei, Huawei and Yang, Peiji and Wang, Zhisheng},
  booktitle={Proceedings of the AAAI Conference on Artificial Intelligence},
  year={2024}
}

@inproceedings{fu2024mimic,
  title={Mimic: Speaking style disentanglement for speech-driven 3d facial animation},
  author={Fu, Hui and Wang, Zeqing and Gong, Ke and Wang, Keze and Chen, Tianshui and Li, Haojie and Zeng, Haifeng and Kang, Wenxiong},
  booktitle={Proceedings of the AAAI conference on artificial intelligence},
  volume={38},
  number={2},
  pages={1770--1777},
  year={2024}
}

@article{liu2025identity,
  title={Identity-Preserving Video Dubbing Using Motion Warping},
  author={Liu, Runzhen and Lin, Qinjie and Liu, Yunfei and Lin, Lijian and Zhu, Ye and Li, Yu and Xian, Chuhua and Hong, Fa-Ting},
  journal={arXiv preprint arXiv:2501.04586},
  year={2025}
}

@article{zhang2024musetalk,
  title={MuseTalk: Real-Time High-Fidelity Video Dubbing via Spatio-Temporal Sampling},
  author={Zhang, Yue and Zhong, Zhizhou and Liu, Minhao and Chen, Zhaokang and Wu, Bin and Zeng, Yubin and Zhan, Chao and He, Yingjie and Huang, Junxin and Zhou, Wenjiang},
  journal={arXiv preprint arXiv:2410.10122},
  year={2024}
}

@inproceedings{chen2025disenemo,
  title={DisenEmo: Learning disentangled emotional representation from facial motion for 3D talking head generation},
  author={Chen, Ziang and Qi, Tianhua and Lu, Cheng and Zheng, Wenming},
  booktitle={2025 IEEE International Conference on Image Processing (ICIP)},
  pages={289--294},
  year={2025},
  organization={IEEE}
}

@inproceedings{josi2024serep,
  title={SEREP: Semantic Facial Expression Representation for Robust In-the-Wild Capture and Retargeting},
  author={Josi, Arthur and Hafemann, Luiz Gustavo and Dib, Abdallah and Got, Emeline and Cruz, Rafael MO and Carbonneau, Marc-Andre},
  booktitle={Proceedings of the IEEE/CVF International Conference on Computer Vision},
  pages={14538--14548},
  year={2025}
}

@inproceedings{xu2024kmtalk,
  title={Kmtalk: Speech-driven 3d facial animation with key motion embedding},
  author={Xu, Zhihao and Gong, Shengjie and Tang, Jiapeng and Liang, Lingyu and Huang, Yining and Li, Haojie and Huang, Shuangping},
  booktitle={European Conference on Computer Vision},
  pages={236--253},
  year={2024},
  organization={Springer}
}

@inproceedings{xing2023codetalker,
  title={Codetalker: Speech-driven 3d facial animation with discrete motion prior},
  author={Xing, Jinbo and Xia, Menghan and Zhang, Yuechen and Cun, Xiaodong and Wang, Jue and Wong, Tien-Tsin},
  booktitle={Proceedings of the IEEE/CVF Conference on Computer Vision and Pattern Recognition},
  pages={12780--12790},
  year={2023}
}

@inproceedings{chen2025cafe,
  title={Cafe-talk: Generating 3d talking face animation with multimodal coarse-and fine-grained control},
  author={Chen, Hejia and Zhang, Haoxian and Zhang, Shoulong and Liu, Xiaoqiang and Zhuang, Sisi and Wan, Pengfei and ZHANG, Di and Li, Shuai},
  booktitle={International Conference on Learning Representations},
  year={2025}
}

@article{barrault2023seamlessw2vbert2,
  title={Seamless: Multilingual Expressive and Streaming Speech Translation},
  author={Barrault, Lo{\"\i}c and Chung, Yu-An and Meglioli, Mariano Coria and Dale, David and Dong, Ning and Duppenthaler, Mark and Duquenne, Paul-Ambroise and Ellis, Brian and Elsahar, Hady and Haaheim, Justin and others},
  journal={arXiv preprint arXiv:2312.05187},
  year={2023}
}

@InProceedings{Cudeiro_2019_CVPRVOCA,
author = {Cudeiro, Daniel and Bolkart, Timo and Laidlaw, Cassidy and Ranjan, Anurag and Black, Michael J.},
title = {Capture, Learning, and Synthesis of 3D Speaking Styles},
booktitle = {Proceedings of the IEEE/CVF Conference on Computer Vision and Pattern Recognition (CVPR)},
month = {June},
year = {2019}
}

@inproceedings{liu2024emoface,
  title={Emoface: Audio-driven emotional 3d face animation},
  author={Liu, Chang and Lin, Qunfen and Zeng, Zijiao and Pan, Ye},
  booktitle={2024 IEEE Conference Virtual Reality and 3D User Interfaces (VR)},
  pages={387--397},
  year={2024},
  organization={IEEE}
}

@article{bigioi2022pose,
  title={Pose-aware speech driven facial landmark animation pipeline for automated dubbing},
  author={Bigioi, Dan and Jordan, Hugh and Jain, Rishabh and McDonnell, Rachel and Corcoran, Peter},
  journal={IEEE Access},
  volume={10},
  pages={133357--133369},
  year={2022},
  publisher={IEEE}
}

@article{bigioi2023multilingual,
  title={Multilingual video dubbing—a technology review and current challenges},
  author={Bigioi, Dan and Corcoran, Peter},
  journal={Frontiers in signal processing},
  volume={3},
  pages={1230755},
  year={2023},
  publisher={Frontiers Media SA}
}

@article{sagonas2016300,
  title={300 faces in-the-wild challenge: Database and results},
  author={Sagonas, Christos and Antonakos, Epameinondas and Tzimiropoulos, Georgios and Zafeiriou, Stefanos and Pantic, Maja},
  journal={Image and vision computing},
  volume={47},
  pages={3--18},
  year={2016},
  publisher={Elsevier}
}

@inproceedings{lipman2023flow,
  title={Flow Matching for Generative Modeling},
  author={Lipman, Yaron and Chen, Ricky TQ and Ben-Hamu, Heli and Nickel, Maximilian and Le, Matt},
  booktitle={11th International Conference on Learning Representations, ICLR 2023},
  year={2023}
}

@article{lipman2024flow,
  title={Flow matching guide and code},
  author={Lipman, Yaron and Havasi, Marton and Holderrieth, Peter and Shaul, Neta and Le, Matt and Karrer, Brian and Chen, Ricky TQ and Lopez-Paz, David and Ben-Hamu, Heli and Gat, Itai},
  journal={arXiv preprint arXiv:2412.06264},
  year={2024}
}

@inproceedings{liuflow,
  title={Flow Straight and Fast: Learning to Generate and Transfer Data with Rectified Flow},
  author={Liu, Xingchao and Gong, Chengyue and others},
  booktitle={NeurIPS 2022 Workshop on Score-Based Methods},
  year={2022}
}

@inproceedings{min2021meta,
  title={Meta-stylespeech: Multi-speaker adaptive text-to-speech generation},
  author={Min, Dongchan and Lee, Dong Bok and Yang, Eunho and Hwang, Sung Ju},
  booktitle={International Conference on Machine Learning},
  pages={7748--7759},
  year={2021},
  organization={PMLR}
}

@article{series2014method,
  title={Method for the subjective assessment of intermediate quality level of audio systems},
  author={Series, B},
  journal={International Telecommunication Union Radiocommunication Assembly},
  volume={2},
  year={2014}
}

@inproceedings{pan2022vocal,
  title={Vocal: Vowel and consonant layering for expressive animator-centric singing animation},
  author={Pan, Yifang and Landreth, Chris and Fiume, Eugene and Singh, Karan},
  booktitle={SIGGRAPH Asia 2022 Conference Papers},
  pages={1--9},
  year={2022}
}

@article{edwards2016jali,
  title={Jali: an animator-centric viseme model for expressive lip synchronization},
  author={Edwards, Pif and Landreth, Chris and Fiume, Eugene and Singh, Karan},
  journal={ACM Transactions on graphics (TOG)},
  volume={35},
  number={4},
  pages={1--11},
  year={2016},
  publisher={ACM New York, NY, USA}
}

@article{zhou2018visemenet,
  title={Visemenet: Audio-driven animator-centric speech animation},
  author={Zhou, Yang and Xu, Zhan and Landreth, Chris and Kalogerakis, Evangelos and Maji, Subhransu and Singh, Karan},
  journal={ACM Transactions on Graphics (ToG)},
  volume={37},
  number={4},
  pages={1--10},
  year={2018},
  publisher={ACM New York, NY, USA}
}

@inproceedings{li2025towardsdynamictexures,
  title={Towards high-fidelity 3d talking avatar with personalized dynamic texture},
  author={Li, Xuanchen and Wang, Jianyu and Cheng, Yuhao and Zeng, Yikun and Ren, Xingyu and Zhu, Wenhan and Zhao, Weiming and Yan, Yichao},
  booktitle={Proceedings of the Computer Vision and Pattern Recognition Conference},
  pages={204--214},
  year={2025}
}

@inproceedings{chu2025artalk,
  title={Artalk: Speech-driven 3d head animation via autoregressive model},
  author={Chu, Xuangeng and Goswami, Nabarun and Cui, Ziteng and Wang, Hanqin and Harada, Tatsuya},
  booktitle={Proceedings of the SIGGRAPH Asia 2025 Conference Papers},
  pages={1--9},
  year={2025}
}

@article{chai2025semantic,
  title={A Semantic Talking Style Space for Speech-Driven Facial Animation},
  author={Chai, Yujin and Weng, Yanlin and Shao, Tianjia and Zhou, Kun},
  journal={IEEE Transactions on Visualization and Computer Graphics},
  year={2025},
  publisher={IEEE}
}
}

\appendix
\section*{Appendices} 
\section{Data collection protocol}\label{sup:data_collection_protocol}

We built our dataset by recording synchronized speech audio and high-fidelity facial performances from 12 professional actors. To ensure broad demographic coverage, we conducted a targeted casting aimed at maximizing diversity in age, gender, and ethnicity within the limited number of actors we captured. Our final cast includes six male and six female actors, with four identifying as White, three as Asian, three as Black, one as Middle Eastern, and one as Latin, spanning an age range from 18 to 64.

Data was captured in two configurations: (1) a seated light-stage setup consisting of six synchronized cameras, and (2) a head-mounted rig containing four cameras aimed at the actor's face.

Actors performed a curated set of actor-specific sentences across eight target emotional states: Neutral, Angry, Fear, Happy, Disgust, Sad, Surprise, and Authority. Except for Neutral, each emotion was recorded at two intensity levels: Moderate and Heightened. The Moderate condition corresponds to a controlled, conversational delivery appropriate for everyday dialogue, whereas the Heightened condition reflects amplified facial expressiveness suited for dramatic or high-impact narrative moments commonly found in digital media. All the performances were conducted in English. 

Because actor expressiveness was critical to the overall quality of the dataset, we designed emotion-specific sentences to help actors deliver consistent, contextually coherent performances. Example sentences associated with each emotion are provided in Table \ref{tab:example_sentences}.

\begin{table}[h!]
\centering
\begin{tabular}{|l|p{0.65\columnwidth}|}
\hline
\textbf{Emotion} & \textbf{Example Sentence} \\ \hline
 Angry & I'm not telling you again. You owe me money, and you need to pay up. \\ \hline
 Fear & I don't like the look of that building, let's try a different route. \\ \hline
 Happy & Do you think this is a good day for a picnic? \\ \hline
 Disgust & Did you see the state of their bathroom? I couldn't even use it. \\ \hline
 Sad & The world keeps turning, but I feel like I'm stuck in a still frame. \\ \hline
 Surprise & <gasping> How is this even possible?! Everything we thought we knew is wrong! \\ \hline
 Authority & You there, stop what you're doing and come over here immediately.\\ \hline
\end{tabular}
\caption{Example sentences with the emotion to be portrayed.}
\label{tab:example_sentences}
\end{table}

Following capture, we followed the standard cinematic‑quality procedure of the Faceform 4D‑capture system\footnote{\href{https://docs.faceform.com/Wrap/4DProcessingPipeline/4DProcessingPipeline.html}{Faceform 4D processing pipeline}}. More precisely, all footage was processed through a standard photogrammetry pipeline, in which each video frame was 3D-reconstructed using RealityCapture, a commercial photogrammetry software. Then, raw mesh sequences were aligned to a common mesh topology of 13,473 vertices using Faceform Wrap 4D, with additional manual refinements by an experienced facial motion-capture supervisor to achieve accurate and stable results. Processing a single minute of footage under this full pipeline takes approximately 8 hours.

\section{Additional implementation details}\label{sup:additional_implementation_details}

\subsection{Training details}\label{sup:training_details}
\noindent \textbf{Importance weights.} The following importance weights are used in eq. 4 of the main manuscript: $\lambda_\text{mse}=100$, $\lambda_\text{vel}=100$, and $\lambda_\text{KL} = 1e-6$.

\noindent \textbf{Optimization hyperparameters.} We trained with a batch size of 32, a learning rate of $1e-4$, and a weight decay of $1e-06$.

\noindent \textbf{Temporal-style dropout.} We randomly drop the temporal style conditioning signal with a probability of 0.1 during training, analogous to classifier-free guidance strategies. This encourages the model to rely on global style for full-face coherence when temporal information is absent. We experimented with higher dropout rates (0.2–0.3) and observed no significant qualitative differences, leading us to adopt 0.1 as a stable default.

\begin{table*}[!t]
    \centering
    \small
    \begin{tabular}{|l|p{12.5cm}|}
        \hline
        \textbf{Function} & \textbf{Details} \\ \hline
        
        Audio encoder & 
        \begin{itemize} 
            \item \textbf{w2v-BERT 2.0()}
            \item \textbf{Linear:} 
            \texttt{Linear(1024$\to$512)}
        \end{itemize} \\ \hline

        Global style encoder &  
        \begin{itemize}
            \item \textbf{Input Layers:} 
                \begin{itemize}
                    \item \texttt{Conv1d(165$\to$512,k=3,p=1)} $\to$ \texttt{Dropout(p=0.2)} $\to$ \texttt{ELU} $\to$ \texttt{LayerNorm(512)}
                    \item \texttt{Conv1d(512$\to$512,k=3,p=1)} $\to$ \texttt{Dropout(p=0.2)} $\to$ \texttt{ELU} $\to$ \texttt{LayerNorm(512)}
                \end{itemize}
            \item \textbf{Positional Encoding:} \texttt{Dropout(p=0.1)}
            \item \textbf{Transformer Encoder Layer:} 
                \begin{itemize}
                    \item \texttt{MultiheadAttention(512)} $\to$ \texttt{Dropout(p=0.1)}
                    \item \texttt{FeedForward: Linear(512$\to$512)} $\to$ \texttt{Dropout(p=0.1)} $\to$ \texttt{Linear(512$\to$512)}
                    \item \texttt{LayerNorm(512)} (x2)
                \end{itemize}
            \item \textbf{Output Layers:} 
                \begin{itemize}
                    \item \texttt{Conv1d(512$\to$1024,k=3,p=1)} $\to$ \texttt{Dropout(p=0.1)} $\to$ \texttt{ELU} $\to$ \texttt{LayerNorm(1024)}
                    \item \texttt{Conv1d(1024$\to$1024,k=3,p=1)}
                \end{itemize}
        \end{itemize} \\ \hline
        
        Temporal style encoder &  
        \begin{itemize}
            \item \textbf{MLP:} \texttt{Linear(141$\to$512)} $\to$ \texttt{GELU} $\to$ \texttt{Linear(512$\to$512)}
        \end{itemize} \\ \hline

 Flow transformer model  &
       \begin{itemize}
  \item \textbf{TransformerLayers (×8):} 
    \begin{itemize}
      \item \textbf{SelfAttention:}
        \begin{itemize}
          \item \textbf{Projections:} \texttt{Linear(512$\to$512)} for $q$, $k$, $v$, and output
          \item \textbf{Dropout:} \texttt{Dropout(p=0.1)}
          \item \textbf{Norm:} \textbf{AdaLN} with conditioning net: \texttt{Linear(512$\to$512)} $\to$ \texttt{SiLU} $\to$ \texttt{Linear(512$\to$1024)}
        \end{itemize}
      \item \textbf{CrossAttention:}
        \begin{itemize}
          \item \textbf{Projections:} \texttt{Linear(512$\to$512)} for $q$, $k$, $v$, and output
          \item \textbf{Dropout:} \texttt{Dropout(p=0.1)}
          \item \textbf{Norm:} \textbf{AdaLN} with conditioning net: \texttt{Linear(512$\to$512)} $\to$ \texttt{SiLU} $\to$ \texttt{Linear(512$\to$1024)}
        \end{itemize}
      \item \textbf{FeedForward:} \texttt{Linear(512$\to$1024)} $\to$ \texttt{Dropout(p=0.1)} $\to$ \texttt{Linear(1024$\to$512)}
      \item \textbf{Norm (FF):} \textbf{AdaLN} with conditioning net: \texttt{Linear(512$\to$512)} $\to$ \texttt{SiLU} $\to$ \texttt{Linear(512$\to$1024)}
    \end{itemize}
\end{itemize}
\begin{itemize}
  \item \textbf{Head MLP:} \texttt{Linear(512$\to$256)} $\to$ \texttt{GELU} $\to$ \texttt{Linear(256$\to$128)}
\end{itemize}
\\ \hline
    \end{tabular}
    \caption{\normalsize Full model architecture, including the audio encoder, global style encoder, temporal style encoder, and flow transformer model.}
    \label{tab:model_detailed_desc}
\end{table*}

\subsection{Model and Representations}\label{sup:model_and_representations}

\noindent \textbf{Model architecture.} We detail our model architecture in \cref{tab:model_detailed_desc}, including audio encoder, global style encoder, temporal style encoder, and our flow transformer model. 

\noindent \textbf{PCA construction and dimensionality.} We construct the PCA basis from the full training set of eight identities using vertex displacements computed with respect to each identity’s neutral face mesh in rest pose. We retain $l=128$ principal components, which preserve approximately 99.5\% of the total variance. This dimensionality provides a good balance between compression and reconstruction fidelity, while enabling efficient sequence modeling. We did not further reduce $l$, as maintaining high-fidelity facial motion is critical for our target applications.

\noindent \textbf{Keypoint signal.} We use a fixed set of $K=55$ full-face keypoints and $K_\text{u}=47$ upper-face keypoints. The upper-face set is a subset of the full-face set, as illustrated in Fig.6 of the main paper.

\noindent \textbf{Keypoint extraction and consistency.} Since all meshes share a fixed topology (see supplementary section A), keypoints are defined via direct vertex correspondences rather than learned detectors. This ensures consistent spatial alignment across identities and sequences.

\noindent \textbf{Keypoint preprocessing.} Keypoint signals are computed as vertex displacements with respect to each identity’s neutral face mesh in rest pose. These displacements are standardized per keypoint using training-set statistics before being fed to the model. This consistent parameterization ensures robustness across identities while improving conditioning stability and generalization. 

\subsection{Facial region masks}\label{sup:facial_region_masks}
The facial masks $\mathcal{M}_{\text{lower}}$ and $\mathcal{M}_{\text{upper}}$ are vectors of length $13{,}473$, with all elements in the range $[0,1]$.
To construct those, we first define two disjoint vertex sets on the template mesh: a lower-face region $\mathbf{R}_{\text{lower}}$ and an upper-face region $\mathbf{R}_{\text{upper}}$, leaving a middle band of vertices unassigned to either region to enable a smooth integration of lower-and upper-face.

To construct $\mathcal{M}_{\text{lower}}$, which emphasizes audio-driven deformations on the lower part of the face, we set:

\begin{equation}
\mathcal{M}_{\text{lower}}(v) =
\begin{cases}
1, & \text{if } v \in \mathbf{R}_{\text{lower}},\\[3pt]
0, & \text{if } v \in \mathbf{R}_{\text{upper}}.
\end{cases}
\end{equation}

For vertices $v$ in the middle band, we define a smooth transition using a distance-based inverse decay from the lower region. We define a point-to-region distance operator $d$ based on Euclidean distance:   
\begin{equation}
d(v,\mathbf{R}_{\text{lower}}) = \min_{u \in \mathbf{R}_{\text{lower}}} \| v - u \|_2,
\end{equation}
and define a lower-face mask as: 
\begin{equation}
\mathcal{M}_{\text{lower}}(v) = \frac{1}{1 + \beta \, d(v,\mathbf{R}_{\text{lower}})}, \quad \text{for } v \notin \mathbf{R}_{\text{lower}} \cup \mathbf{R}_{\text{upper}},
\end{equation}
where $\beta > 0$ controls the rate of inverse decay (we use $\beta = 1.0$).

The upper-face mask is then defined as the complement:
\begin{equation}
\mathcal{M}_{\text{upper}} = 1 - \mathcal{M}_{\text{lower}},
\end{equation}
which, analogously, enforces temporal-style keypoints to drive the upper face.

\subsection{Topology conversion}\label{sup:topology_conversion} To convert mesh data from a source topology to a target topology, we compute a sparse conversion matrix that maps each vertex of the target mesh to a barycentric combination of the closest triangle in the source mesh. Meshes must be spatially aligned and scaled before computing the matrix. The process is as follows: (1) For each vertex in the target topology, we find the closest triangle in the source topology using a point-to-triangle distance metric. (2) We compute the barycentric coordinates of the target vertex with respect to this closest triangle. (3) These barycentric weights are used to populate a sparse matrix, where each row corresponds to a target vertex and each column to a source vertex.

The resulting matrix enables efficient conversion: multiplying the source vertex data by it yields the corresponding data in the target topology.
This method ensures that geometric information is transferred as accurately as possible between different mesh topologies, while the conversion is performed via a single matrix multiplication.



\subsection{Auxiliary Components for Rendering}\label{sup:auxiliary_components_for_rendering}

\noindent \textbf{Textures.} We apply static per-actor textures to the animated meshes to enhance visual realism. Modeling dynamic or view-dependent textures could further improve fidelity, as highlighted by Li \textit{et al.} \cite{li2025towardsdynamictexures}, but such aspects are not the focus of this work. We leave these extensions to future research.

\begin{figure}
    \centering
    \includegraphics[width=\linewidth]{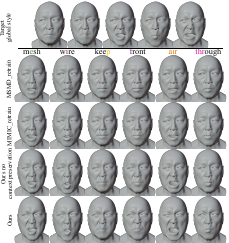}
    \caption{Raw geometry renderings for the different baselines with a desired angry target style sequence while the audio intent mostly neutral.}
    \label{fig:example_based_main_v2_raw}
\end{figure}

\begin{figure*}
    \centering
    \includegraphics[width=\linewidth]{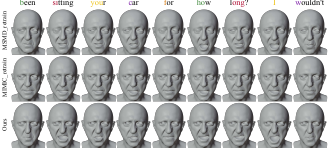}
    \caption{Raw geometry renderings for qualitative comparison of MSMD\_retrain, MIMIC\_retrain, and Ours in the matching context scenario, where the target style matches the audio intent. We do not display the target to encourage focus on the lip-sync and naturalness.}
    \label{fig:previous_window_raw}
\end{figure*}

\noindent \textbf{Eye and jaw processing.}
For all retrained models, we predict eye motion and jaw articulation using lightweight regressors trained directly on top of the model's latent motion features. Given the predicted motion feature sequence $\hat{\mathbf{x}} \in \mathbb{R}^{N \times l}$, where $N$ denotes the number of frames and $l$ the feature dimension, we train a small MLP to infer (i) eye gaze and (ii) jaw rotation.

Eye motion is represented by four parameters corresponding to yaw and pitch for the left and right eyes, while jaw articulation is represented as a $3 \times 3$ rotation matrix parameterized as a 9-dimensional vector.

For each target sequence $y_{\text{eyes\_jaw}} \in \mathbb{R}^{N \times 13}$, the MLP predicts $\hat{y}_\text{eyes\_jaw} = g(\hat{\mathbf{x}})$ and is optimized using ${L_{\text{eyes\_jaw}}}$ defined with a combination of mean-squared error (MSE) and velocity consistency losses:
\begin{equation}
\begin{aligned}
\mathcal{L}_{\text{eyes\_jaw}} =
&\sum_{n=0}^{N} \left\| \hat{y}_{\text{eyes\_jaw}, n} - y_{\text{eyes\_jaw}, n} \right\|_2^2 \\
&+ \sum_{n=1}^{N} \left\| 
\left(\hat{y}_{\text{eyes\_jaw}, n} - \hat{y}_{\text{eyes\_jaw}, n-1}\right) \right. \\
&\qquad\left. -
\left(y_{\text{eyes\_jaw}, n} - y_{\text{eyes\_jaw}, n-1}\right)
\right\|_2^2.
\end{aligned}
\end{equation}

\noindent where both terms are weighted equally.

This post-hoc regression strategy enables satisfying estimation of eye and jaw motion \textbf{while keeping the core facial-motion model unchanged}. We emphasize that these components are included primarily to enhance visual realism rather than as a core contribution of our method.

While such components could be integrated directly into the mesh topology for end-to-end prediction, most production pipelines treat eyes and teeth as separate assets and animate them independently to meet scene-specific requirements. Our design, therefore, aligns with common practices while maintaining architectural simplicity.


\section{Raw geometry results}\label{sup:raw_geometry_results}

This section presents additional qualitative results showing facial geometry only (without textures, eyes, or teeth). These visualizations correspond directly to the model’s predictions, while rendering assets are used in the main paper solely to improve perceptual realism.

Fig.~\ref{fig:example_based_main_v2_raw} presents additional geometry-only results for the example-based style generation task, where neutral audio is paired with an angry target style, similar to fig.~4 of the main paper. Fig.~\ref{fig:previous_window_raw} presents results for the lip-synchronization experiment, similar to fig.~7 in the main paper. The same observations apply.

\begin{figure*}[t]
    \centering
    \includegraphics[width=0.95\linewidth]{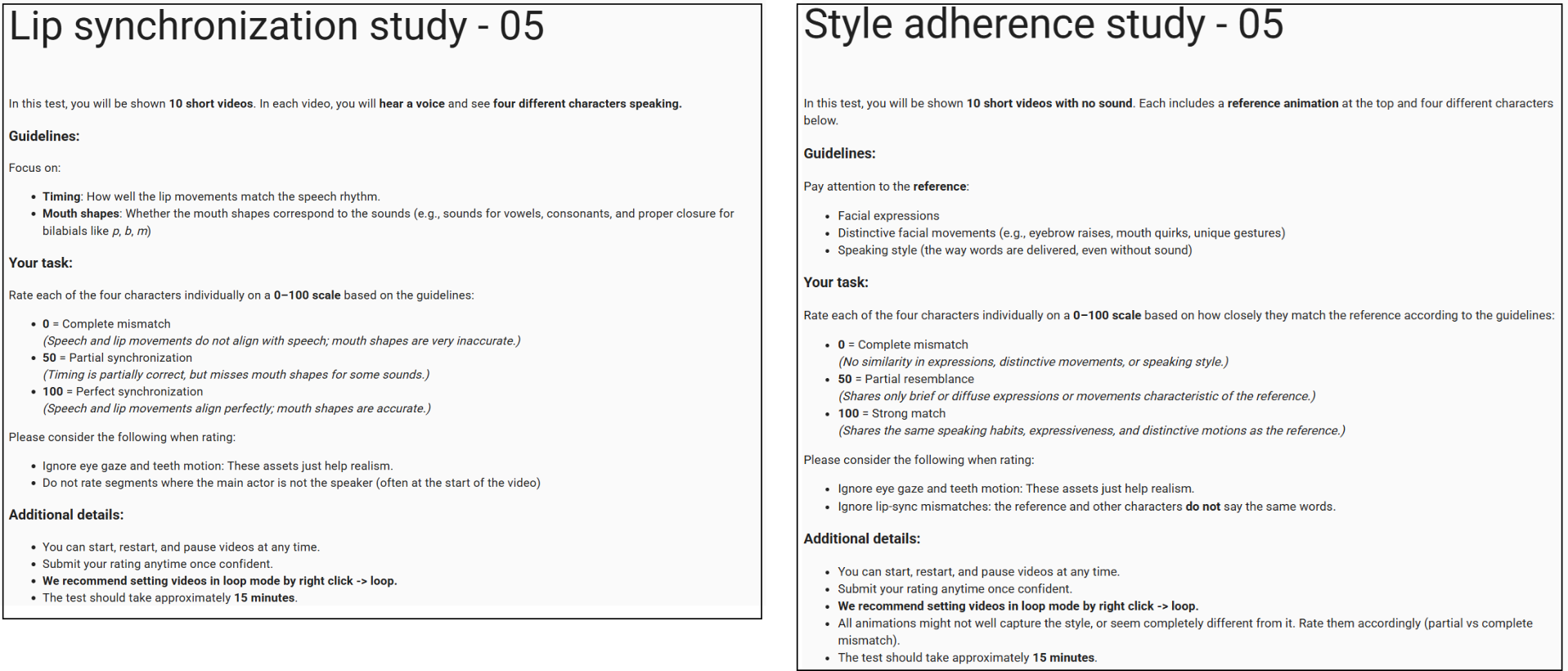}
    \caption{Guidelines for the lip-synchronization (left) and style-adherence (right) user studies.}
    \label{fig:guidelines}
    \vspace{0.5em}
    
    \includegraphics[width=0.95\linewidth]{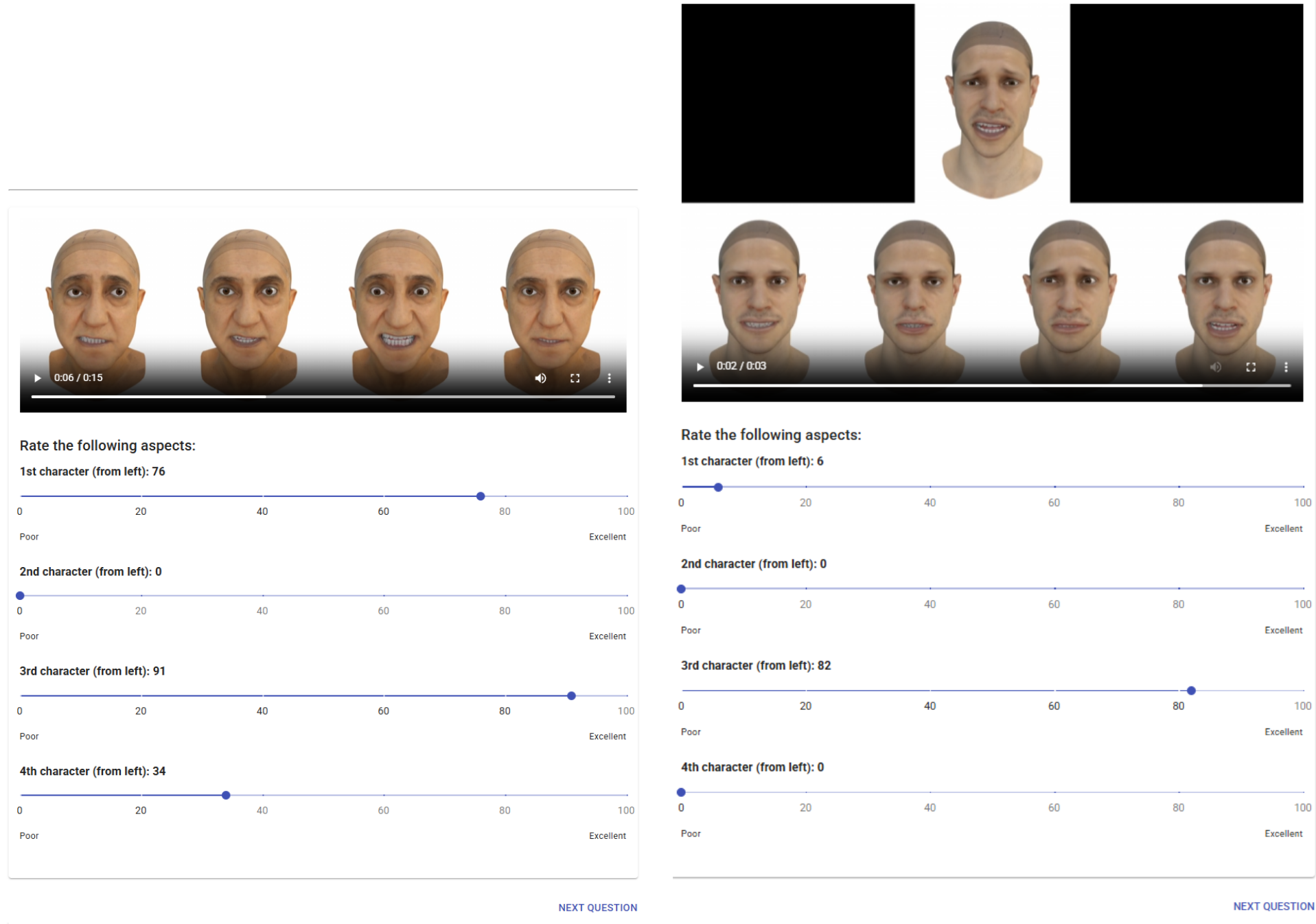}
    \caption{Videos and rating interface for the lip-synchronization (left) and style-adherence (right) studies.}
    \label{fig:example_videos}    
\end{figure*}

\section{Perceptual study}\label{sup:perceptual_study}

\subsection{Perceptual study construction details}

We present screenshots of the user‑study guidelines for both the lip‑synchronization and the style‑adherence in  \cref{fig:guidelines}. For the lip\allowbreak‑synchronization study, raters were shown videos containing the four model outputs, displayed in random order, and asked to rate each generation individually on a scale from 0 to 100. For the style‑adherence study, the videos were muted to ensure raters focused solely on the delivery style, while the target-style animation was displayed in a separate upper row, playing simultaneously with the generated outputs. See \cref{fig:example_videos} for screenshots of both studies videos and rating display. Participants were allowed to replay the videos as many times as needed and encouraged to submit their ratings only when confident in their assessments. They were also asked to ignore eyes and teeth motion at all time, given those assets are added to improve realism and not part of the evaluated methods.  

\noindent \textbf{Negative anchor} For the lip‑synchronization study, the negative anchor is randomly sampled from other available sequences of the same actor.

In the style‑adherence study, constructing a meaningful negative anchor is more challenging, as some emotion categories are semantically and behaviorally close (e.g., surprise vs. happiness, or authority vs. anger). To obtain more robust negatives, we represent each animation clip using summary statistics of its facial landmark dynamics, computed from the KM‑Speaker set of 55 keypoints. For each frame, we concatenate the flattened 3D landmark positions with their first‑order temporal velocities, forming a per‑frame descriptor. For each clip, we then compute the temporal mean and covariance of these descriptors.

To select a negative emotion class for a given target emotion, we compute an inter‑emotion distance based on the centroid statistics of each emotion class. Given a target animation, the negative sample is drawn from the emotion class that maximizes this distance. All centroid statistics and distance computations are performed within the test actor’s data distribution only, as different actors exhibit distinct styles and interpretations when conveying emotions.



\subsection{Perceptual study additional analysis}
\begin{table}[ht]
\centering
\caption{Mean rankings for lip Synchronization and style adherence user studies.}
\label{tab:mean_ranking_merged}
\begin{tabular}{lcc}
\toprule
\textbf{Method}        & \textbf{Lip Sync} & \textbf{Style Adherence} \\
        & (Mean Rank)$\downarrow$ & (Mean Rank)$\downarrow$ \\
\midrule
Negative        & 4.00 & 3.30 \\
MIMIC\_retrain          & 2.90 & 2.69 \\
MSMD\_retrain           & 1.75 & 2.75 \\
Ours                   & \textbf{1.35} & \textbf{1.26} \\
\bottomrule
\end{tabular}
\end{table}

To complete the analysis of the pursued user studies, we additionally compute the mean ranking for each method (\cref{tab:mean_ranking_merged}), including the negative anchor, and ensured statistical significance ($\rho < 0.01$) using a Wilcoxon test between our methods' performances and each baseline and negative.  

Across both user studies, our method consistently achieves the lowest mean ranking, demonstrating its consistency across diverse generations and styles. Importantly, we observe that the mean ranking of the negative in both studies consistently comes behind other methods, while the ordering of MSMD\_retrain and MIMIC\_retrain follows the main paper results. 

We further assess inter-rater agreement to evaluate the reliability of the collected perceptual judgments. For lip synchronization, agreement is strong at the pairwise level, with a mean Pearson correlation of $0.88$, indicating a high degree of consistency in relative comparisons across raters. This level of pairwise agreement suggests that lip synchronization is generally perceived as a more objective task, leading to stable and reproducible evaluations.
For style adherence, we likewise observe consistent inter-rater agreement, with a mean pairwise Pearson correlation of $0.57$. This suggests that, while individual assessments of style may vary due to subjective interpretation, raters nonetheless tend to converge toward a consistent consensus when judgments are considered collectively.

\begin{figure*}
    \centering
    \includegraphics[width=\linewidth]{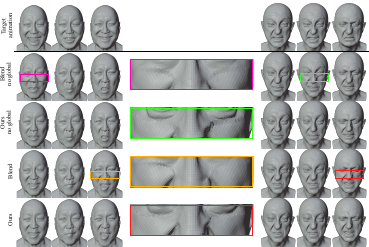}
    \caption{Qualitative results for two distinct examples (left and right), given a target animation (first row), comparing the blending baseline with our model with and without global style conditioning. Zoomed-in visualizations are provided in the center to highlight geometry inconsistencies. Best viewed in color.}
    \label{fig:blending_experiment}
\end{figure*}

\section{Dialogue localization: Blending baseline and global style}\label{sup:blending_baseline}

In this section, we evaluate a blending-based baseline inspired by common practices in artist-driven pipelines, where mouth motion is edited or composed independently from upper-face animation. This setup serves as a strong deterministic reference for dialogue localization.

Specifically, we construct a blended animation by combining target upper-face motion with lip-sync animation generated by our model. The blended animation $\mathcal{X}_{\text{blend}}$ is defined as: 
\begin{equation}
\mathcal{X}_{\text{blend}} =
\mathcal{M}_{\text{upper}} \odot \mathcal{X}_{\text{target}}
+ \mathcal{M}_{\text{lower}} \odot \mathcal{X}_{\text{generated}},
\end{equation}
\noindent where $\odot$ denotes element-wise multiplication, $\mathcal{M}_{\text{upper}}$ and $\mathcal{M}_{\text{lower}}$ are blending masks (Supplementary Section B), and $\mathcal{X}_{\text{target}}$, $\mathcal{X}_{\text{generated}}$ denote the target and generated animations, respectively.

We evaluate this baseline both without global style conditioning (\textit{Blend (no global)}) and with it (\textit{Blend}), to assess its impact on overall coherence.

Qualitative results are shown in \cref{fig:blending_experiment}. While blending preserves upper-face motion, it introduces geometric inconsistencies in transition regions, particularly near mask boundaries (see zoomed-in views), even when global style is applied. These discontinuities affect both surface geometry and normals, and would lead to shading artifacts in rendered animations. 

More fundamentally, such compositing strategies remain sensitive to identity-specific geometry and expression variation. Although careful tuning of blending parameters can improve results in constrained cases, this process is time-consuming and does not generalize well across diverse conditions. Also, while global style conditioning improves coherence between upper- and lower-face motion (particularly visible for the left target style), it does not address the fundamental limitations of compositing-based strategies. In contrast, our model directly generates coherent full-face animations, avoiding boundary artifacts and generalizing across identities without manual intervention. In addition, our approach supports both dialogue localization and example-based generation within a unified framework.



\section{Generalization}\label{sup:generalization}

We conduct a qualitative generalization experiment in the example-based setting to evaluate our model beyond the controlled high-fidelity setting, considering variations in facial geometry, non production quality audio, and styles extracted from off-the-shelf face expression capture method \cite{josi2024serep} rather than 4D capture. To this end, we sample three audio clips from CelebV-HQ \cite{zhu2022celebvhq}, three target videos serving as styles, and three neutral meshes with diverse facial traits from Triplegangers \cite{triplegangers}.

\begin{figure*}
    \centering
    \includegraphics[width=0.7\linewidth]{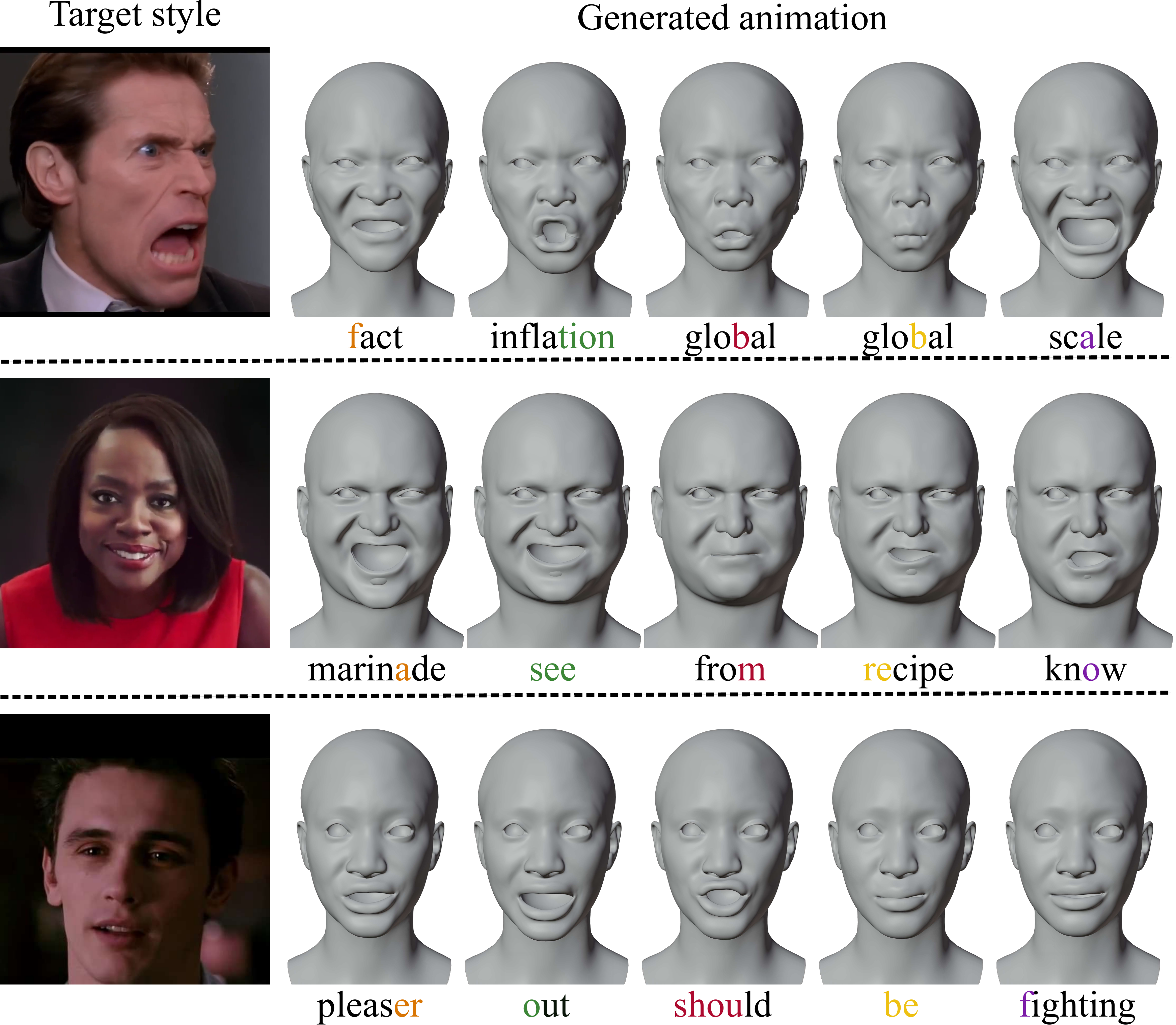}
    \caption{KM-Speaker generalization results on in-the-wild audio, style, and varying face geometries. Given a target video providing the desired style (left), we first extract the corresponding facial expressions onto an arbitrary mesh using an off-the-shelf expression capture model \cite{josi2024serep}. We then generate a speech-driven animation (right) using KM-Speaker, conditioned on the extracted style and a different speech signal. Corresponding dynamic results are shown in the supplementary video (7:20–7:50). Best viewed in color.}
    \label{fig:generalization}
\end{figure*}

We present those qualitative results in \cref{fig:generalization}, with corresponding dynamic results visible in our supplementary video (7:20 to 7:50). Despite these challenging conditions, our model demonstrates reasonable robustness: it captures global style from reconstructed expressions and produces convincing lip motion with accurate synchronization. For instance, expressive styles such as anger (first row) are clearly conveyed through both facial dynamics and motion intensity, whereas more neutral styles (last row) yield correspondingly restrained animations without introducing artifacts.

At the same time, this experiment reveals limitations in geometric generalization. For certain identities, we observe imperfect mouth closure (last row) and inner-mouth mesh intersections (middle row), particularly for challenging facial morphologies. These failure cases highlight the limits of the learned facial prior when applied to out-of-distribution geometries. 

Despite this, the observed artifacts remain moderate. Inner-mouth intersections could likely be mitigated through appropriate regularization, whereas imperfect lip closure remains a more fundamental limitation. In both cases, incorporating identity-specific modeling, such as few-shot adaptation or explicit identity encoding/decoding (e.g., SEREP \cite{josi2024serep} or  \cite{chai2025semantic}), would likely improve robustness and reduce these effects.

\end{document}